\documentclass[runningheads]{llncs}
\usepackage{comment}
\usepackage{url}
\usepackage{amsmath}
\usepackage{amssymb}
\usepackage{adjustbox}
\usepackage{makecell,multirow,booktabs}
\usepackage{xcolor,colortbl}
\usepackage[width=122mm,left=12mm,paperwidth=146mm,height=193mm,top=12mm,paperheight=217mm]{geometry}

\definecolor{our-green}{rgb}{0.56, 0.692, 0.195}
\definecolor{our-darkgreen}{rgb}{0.297, 0.348, 0.105}
\definecolor{our-yellow}{rgb}{0.881, 0.611, 0.142}
\definecolor{our-red}{rgb}{0.923, 0.386, 0.209}

\usepackage{xr}
\usepackage{wrapfig}
\usepackage{wasysym}
\usepackage{graphbox}
\graphicspath{{graphics/}}
\usepackage{hyperref}
\hypersetup{
	colorlinks,
	linkcolor=our-darkgreen,
	urlcolor=our-darkgreen,
	citecolor=our-red}
\usepackage[numbers,sort]{natbib}

\newlength{\fHeight}
\newcommand{\faBicycle}{\includegraphics[height=\fHeight]{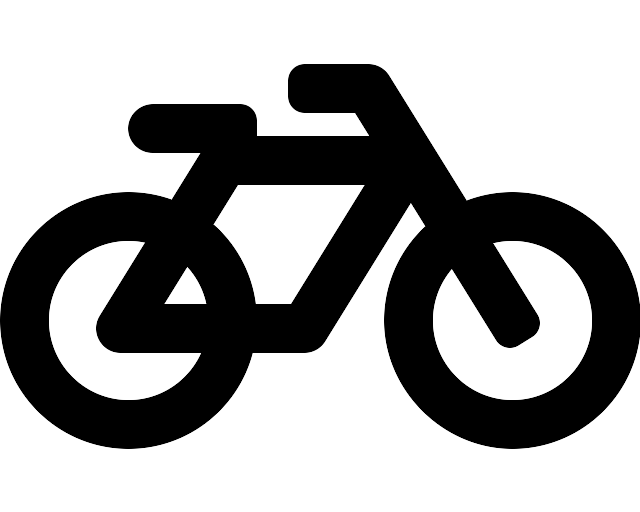}}
\newcommand{\faBus}{\includegraphics[height=\fHeight]{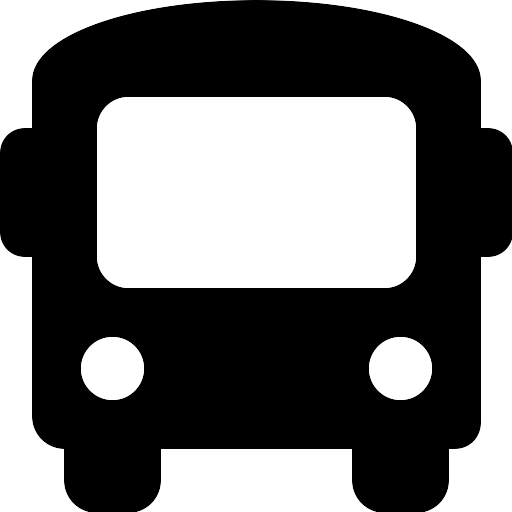}}
\newcommand{\faCar}{\includegraphics[height=\fHeight]{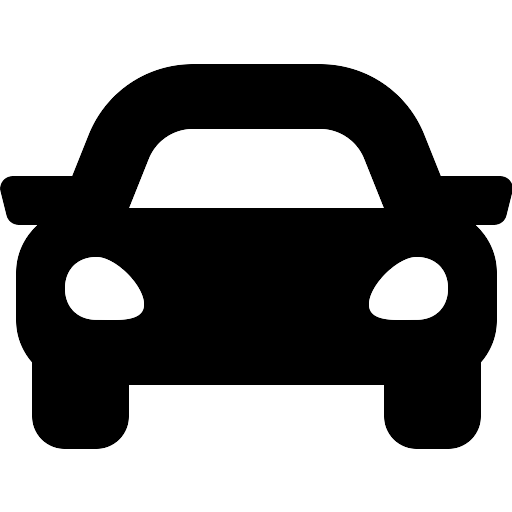}}
\newcommand{\faCat}{\includegraphics[height=\fHeight]{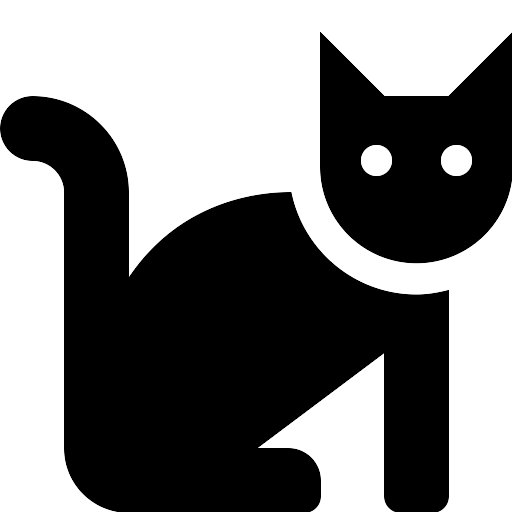}}
\newcommand{\faCow}{\includegraphics[height=\fHeight]{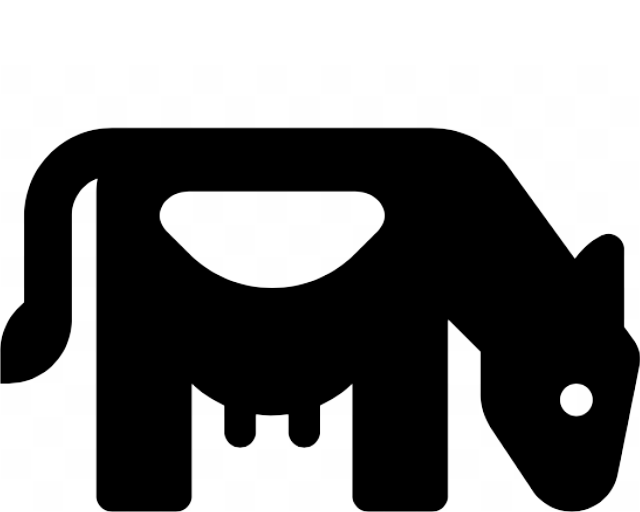}}
\newcommand{\faChair}{\includegraphics[height=\fHeight]{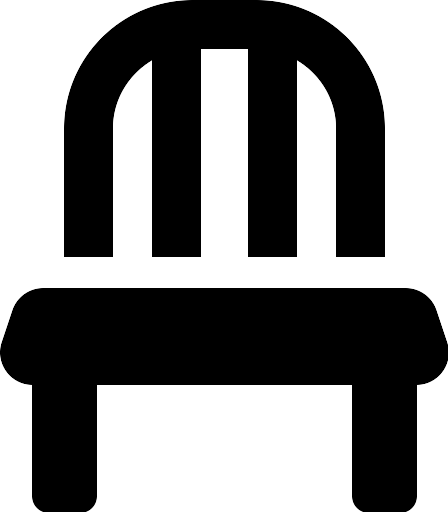}}

\newcommand{\faCrow}{\includegraphics[height=\fHeight]{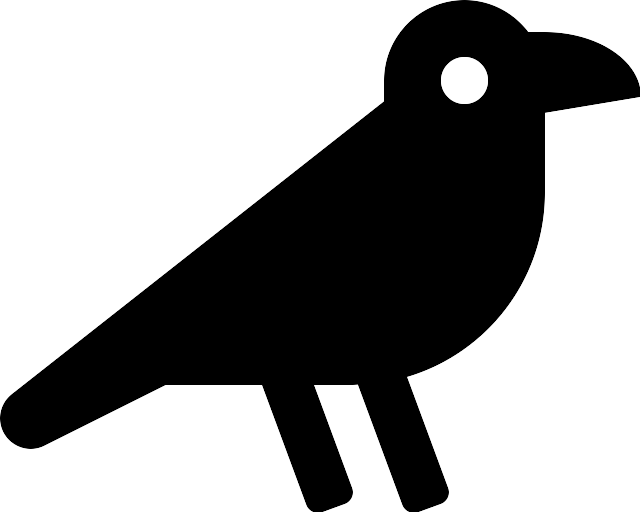}}
\newcommand{\faDog}{\includegraphics[height=\fHeight]{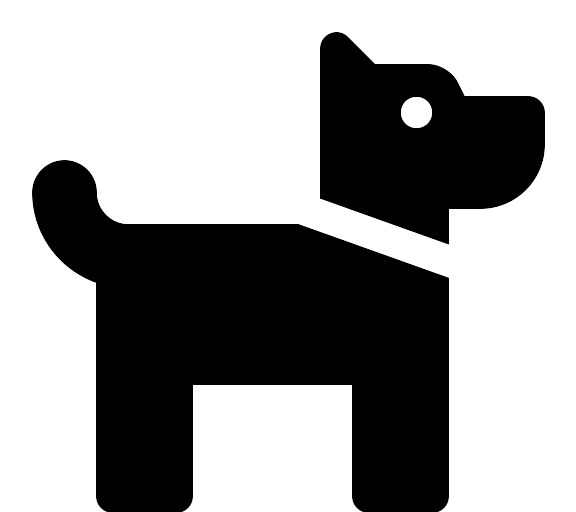}}

\newcommand{\faHorse}{\includegraphics[height=\fHeight]{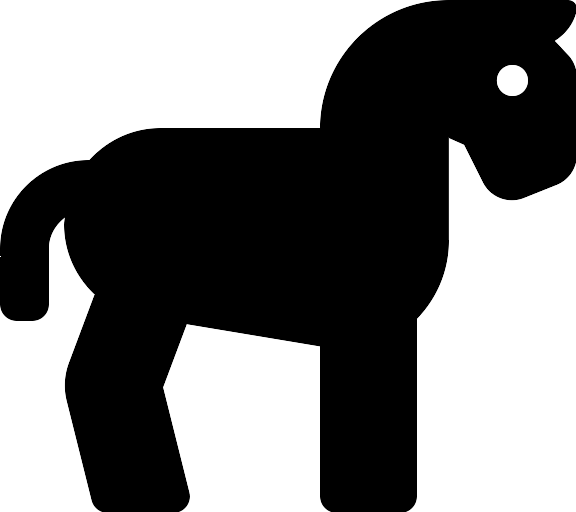}}
\newcommand{\faMotorcycle}{\includegraphics[height=\fHeight]{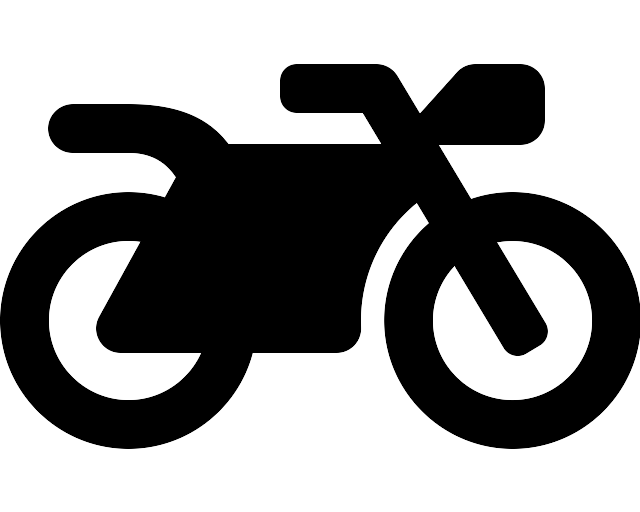}}
\newcommand{\faPlane}{\includegraphics[height=\fHeight]{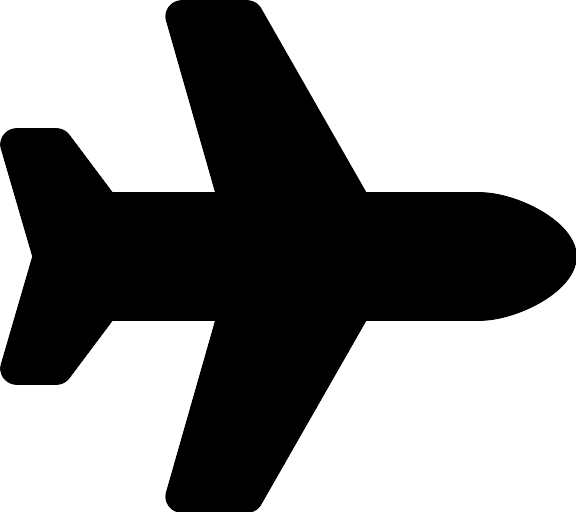}}
\newcommand{\faShip}{\includegraphics[height=\fHeight]{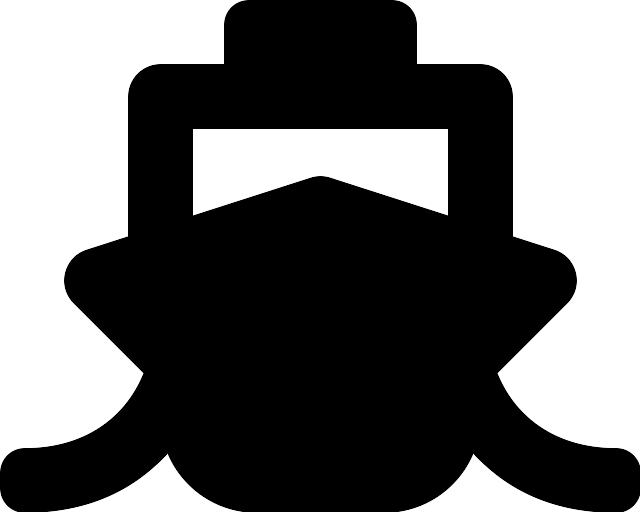}}
\newcommand{\faSheep}{\includegraphics[height=\fHeight]{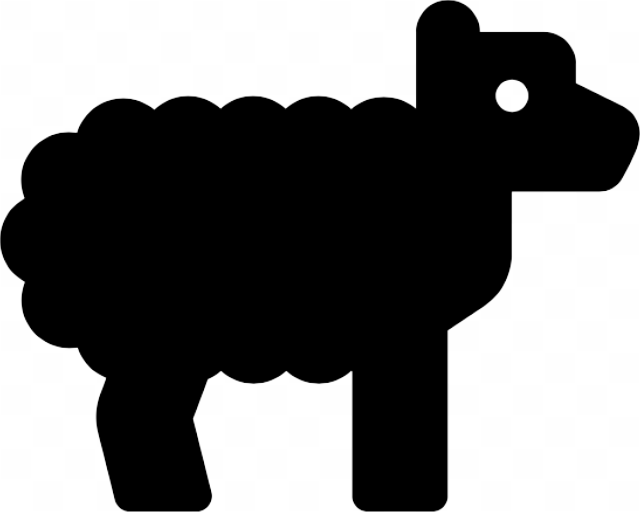}}
\newcommand{\faTrain}{\includegraphics[height=\fHeight]{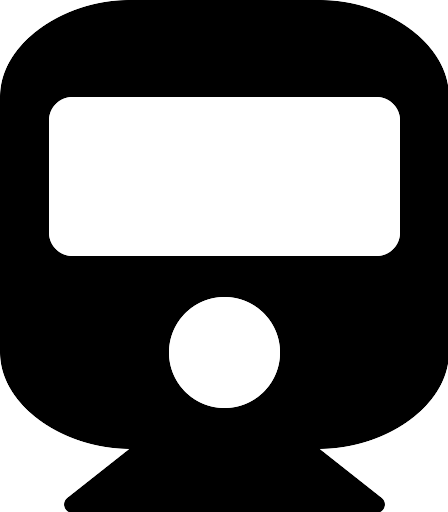}}
\newcommand{\faTulip}{\includegraphics[height=\fHeight]{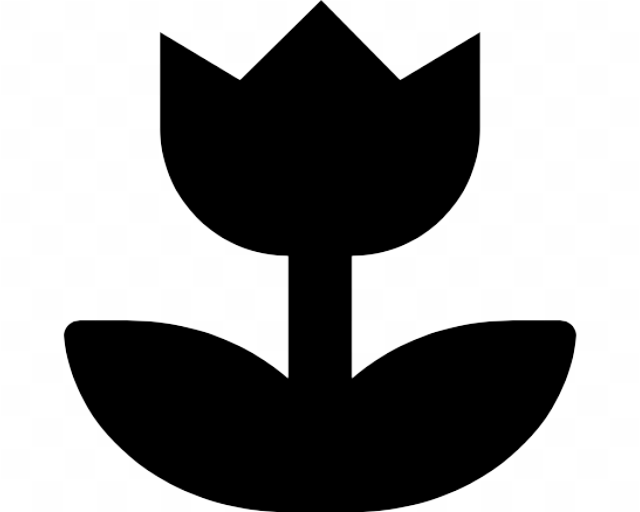}}
\newcommand{\faTv}{\includegraphics[height=\fHeight]{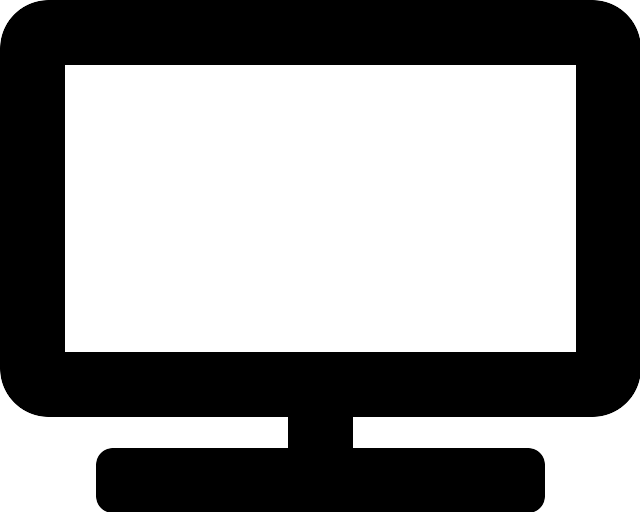}}
\newcommand{\faWalking}{\includegraphics[height=\fHeight]{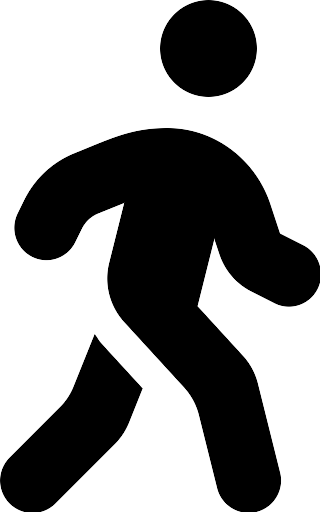}}
\newcommand{\faWineBottle}{\includegraphics[height=\fHeight]{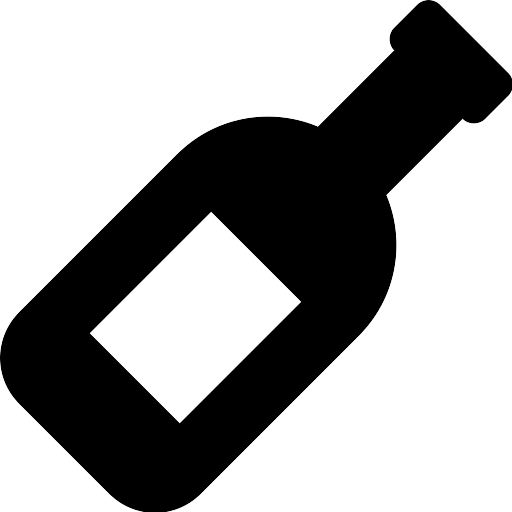}}

\newcommand{\tableIcons}{%
\faPlane & \faBicycle & \faCrow & \faShip & \faWineBottle & \faBus &
\faCar & \faCat &	\faChair & \faCow & \faDinningTable & \faDog & \faHorse &
\faMotorcycle & \faWalking & \faTulip& \faSheep & \faCouch & \faTrain & \faTv}

\usepackage{tabularx}
\usepackage{collcell}
\newcommand\setrow[1]{\gdef\rowmac{#1}\ignorespaces}
\newcommand\clearrow{\global\let\rowmac\relax}
\clearrow

\newcolumntype{C}{>{\collectcell\rowmac}c<{\endcollectcell}}
\newcolumntype{R}{>{\collectcell\rowmac}r<{\endcollectcell}}
\newcolumntype{L}{>{\collectcell\rowmac}l<{\endcollectcell}}

\newcommand{\hlcell}{\cellcolor{black!9}}
\newcommand{\hhlcell}{\cellcolor{black!25}}

\usepackage{caption,subcaption}
\usepackage{algorithm}
\usepackage[noend]{algpseudocode}
\DeclareCaptionSubType*{algorithm}
\algrenewcommand\algorithmicindent{1em}
\algrenewcommand{\algorithmiccomment}[1]{%
\bgroup\hskip2em\textcolor{our-darkgreen}{//~\textsl{#1}}\egroup}
\algrenewcommand{\Return}{\State\textbf{return}\ }
\algnewcommand{\Save}{\State\textbf{save}\ }
\algnewcommand{\Load}{\State\textbf{load}\ }

\usepackage{xspace}
\makeatletter
\DeclareRobustCommand\onedot{\futurelet\@let@token\@onedot}
\def\@onedot{\ifx\@let@token.\else.\null\fi\xspace}

\def\eg{{e.g}\onedot} 
\def\ie{{i.e}\onedot}

\makeatother

\newcommand{\R}{\mathbb{R}}
\newcommand{\ve}{\mathbf{e}}
\newcommand{\vv}{\mathbf{v}}
\DeclareMathOperator*{\argmin}{arg\,min}

\newcommand{\VOC}{Pascal~VOC\xspace}
\newcommand{\SPair}{SPair-71k\xspace}
\newcommand{\WillowOC}{Willow ObjectClass\xspace}
\newcommand{\method}{BB-GM\xspace}

\newcommand{\interKP}{\mbox{$\wasytherefore\cap\wasytherefore$}}
\newcommand{\incluKP}{\mbox{$\wasytherefore\subset\wasytherefore$}}
\newcommand{\unfilKP}{\mbox{$\wasytherefore\cup\wasytherefore$}}

\newcommand{\alg}[1]{Alg.~\ref{alg:#1}}
\newcommand{\fig}[1]{Fig.~\ref{fig:#1}}
\newcommand{\tab}[1]{Tab.~\ref{tab:#1}}
\newcommand{\sect}[1]{Sec.~\ref{sec:#1}}

\begin{document}

\pagestyle{headings}
\mainmatter
\def\ECCVSubNumber{6207}  

\title{Deep Graph Matching via Blackbox Differentiation of Combinatorial Solvers}


\titlerunning{Deep Graph Matching with Combinatorial Solvers}
\author{%
	Michal~Rol\'inek\inst{1}
		\and
	Paul~Swoboda\inst{2}
		\and
	Dominik~Zietlow\inst{1}
		\and\\
	Anselm~Paulus\inst{1}
		\and
	V\'it~Musil\inst{3}
		\and
	Georg~Martius\inst{1}
}
\authorrunning{M.~Rol\'inek et al.}
%
\institute{%
Max Planck Institute for Intelligent Systems, T\"ubingen, Germany
\and
Max Planck Institute for Informatics,	Saarbr\"ucken, Germany
\and
Universit\`a degli Studi di Firenze, Italy\\
\url{github.com/martius-lab/blackbox-deep-graph-matching}\\
\email{michal.rolinek@tue.mpg.de}
}
\maketitle

\begin{abstract}
Building on recent progress at the intersection of combinatorial optimization
and deep learning, we propose an end-to-end trainable architecture for deep
graph matching that contains unmodified combinatorial solvers. Using the
presence of heavily optimized combinatorial solvers together with some
improvements in architecture design, we advance state-of-the-art on deep graph
matching benchmarks for keypoint correspondence.  In addition, we highlight the
conceptual advantages of incorporating solvers into deep learning
architectures, such as the possibility of post-processing with a strong
multi-graph matching solver or the indifference to changes in the training
setting.  Finally, we propose two new challenging experimental setups.
\keywords{
deep graph matching,
keypoint correspondence,
combinatorial optimization
}
\end{abstract}

\section{Introduction}

\begin{wrapfigure}{o}{.37\textwidth}
  \vspace*{-2\baselineskip}
  \captionsetup{justification=centering}
  \setlength{\belowcaptionskip}{-2\baselineskip}
	\includegraphics[width=.365\textwidth]{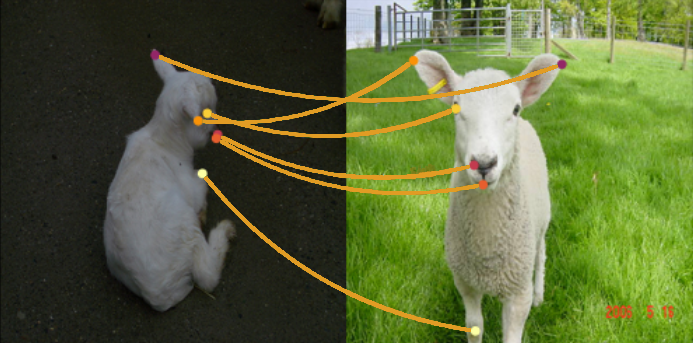}
	\includegraphics[width=.365\textwidth]{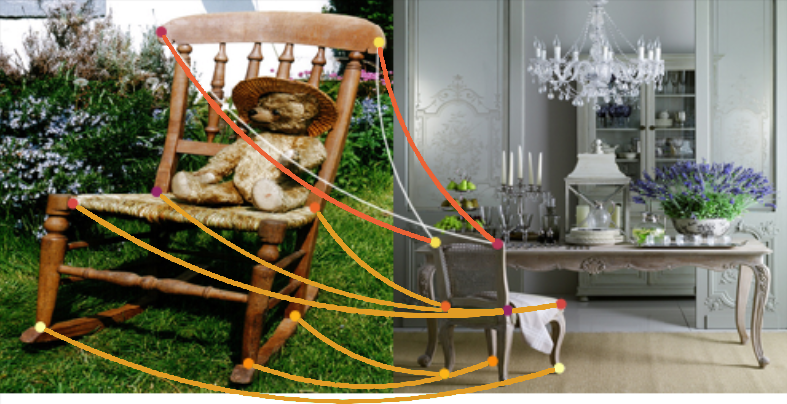}
	\caption{Example keypoint matchings of the proposed architecture on \SPair.}
	\label{fig:examples}
\end{wrapfigure}
Matching discrete structures is a recurring theme in numerous branches of
computer science. Aside from extensive analysis of its theoretical and
algorithmic aspects~\cite{assignment_problems,grohe_et_al:LIPIcs:2018:9602},
there is also a wide range of applications. Computer vision, in particular, is
abundant of tasks with a matching flavor; optical
flow~\cite{Baker:IJCV:11,black_flow,Sun_2018_CVPR}, person
re-identification~\cite{schroff2015facenet,gasse2019exact}, stereo
matching~\cite{Luo2016EfficientDL,chang2018pyramid}, pose
estimation~\cite{Cao_2017_CVPR,gasse2019exact}, object
tracking~\cite{nam2015mdnet,10.1109/ICCV.2015.357}, to name just a few.
Matching problems are also relevant in a variety of scientific disciplines
including biology~\cite{10.1007/978-3-319-10404-1_11}, language
processing~\cite{pmlr-v80-niculae18a},
bioinformatics~\cite{10.1109/TCBB.2015.2474391}, correspondence problems in
computer graphics~\cite{sahilliouglu2019recent} or social network
analysis~\cite{10.5555/3060832.3060869}.

Particularly, in the domain of computer vision, the matching problem has two
parts: \textbf{extraction of local features} from raw images and
\textbf{resolving conflicting evidence} \eg multiple long-term occlusions in a
tracking context.  Each of these parts can be addressed efficiently in
separation, namely by deep networks on the one side and by specialized purely
combinatorial algorithms on the other.  The latter requires a clean abstract
formulation of the combinatorial problem.  Complications arise if concessions
on \emph{either} side harm performance.  Deep networks on their own have a
limited capability of \emph{combinatorial generalization}~\cite{47094} and
purely combinatorial approaches typically rely on fixed features that are often
suboptimal in practice. To address this, many \emph{hybrid} approaches have
been proposed.

In case of \emph{deep graph matching} some approaches rely on finding suitable
differentiable relaxations~\cite{zanfir2018deep,wang2019neural}, while others
benefit from a tailored architecture
design~\cite{wang2019learning,fey2020deep,jiang2019glmnet,Zhang_2019_ICCV}.
What all these approaches have in common is that they compromise on the
combinatorial side in the sense that the resulting ``combinatorial block''
would not be competitive in a purely combinatorial setup.

In this work, we present a novel type of end-to-end architecture for semantic
keypoint matching that \textbf{does not make any concessions on the
combinatorial side} while maintaining strong feature extraction.  We build on
recent progress at the intersection of combinatorial optimization and deep
learning~\cite{VlastelicaEtal2020:BBoxSolvers} that allows to seamlessly embed
\textbf{blackbox implementations} of a wide range of combinatorial algorithms
into deep networks in a \textbf{mathematically sound} fashion.  As a result, we
can leverage heavily optimized graph matching
solvers~\cite{Swoboda2016ASO,swoboda2019convex} based on dual block coordinate
ascent for Lagrange decompositions.

Since the combinatorial aspect is handled by an expert algorithm, we can focus
on the rest of the architecture design: building representative graph matching
instances from visual and geometric information.  In that regard, we leverage
the recent findings~\cite{fey2020deep} that large performance improvement can
be obtained by correctly incorporating relative keypoint locations via
SplineCNN~\cite{fey2018splinecnn}.

Additionally, we observe that correct matching decisions are often simplified
by leveraging global information such as viewpoint, rigidity of the object or
scale (see also \fig{examples}). With this in mind, we propose  a natural
\textbf{global feature attention mechanism} that allows to adjust the weighting
of different node and edge features based on a global feature vector.

Finally, the proposed architecture allows a stronger post-processing step. In
particular, we use a multi-graph matching solver~\cite{swoboda2019convex}
during evaluation to jointly resolve multiple graph matching instances in a
consistent fashion.

On the experimental side, we achieve state-of-the-art results on standard
keypoint matching datasets \VOC (with Berkeley
annotations~\cite{everingham2010pascal,bourdev2009poselets}) and
\WillowOC~\cite{Cho2013:learning}.  Motivated by lack of challenging
standardised benchmarks, we additionally propose two new experimental setups.
The first one is the evaluation on \SPair~\cite{min2019spair71k} a high-quality
dataset that was recently released in the context of \emph{dense image
matching}.  As the second one, we suggest to drop the common practice of
keypoint pre-filtering and as a result force the future methods to address the
presence of keypoints without a match.

The contributions presented in this paper can be summarized as follows.
\begin{enumerate}
\parskip=\medskipamount
	\item We present a novel and conceptually simple \textbf{end-to-end trainable architecture} that seamlessly
	incorporates a state-of-the-art combinatorial graph matching solver. In addition, improvements
	are attained on the feature extraction side by processing global image
	information.
	\item We introduce two new experimental setups and suggest them as future
	benchmarks.
	\item We perform an extensive evaluation on existing benchmarks as well as on
	the newly proposed ones. Our approach reaches higher matching accuracy than
	previous methods, particularly in more challenging scenarios.
	\item We exhibit further advantages of incorporating a combinatorial solver:
	\parskip=0pt
	\begin{enumerate}
	    \item[(i)] possible post-processing with a multi-graph matching solver,
			\item[(ii)] an effortless transition to more challenging scenarios with
			unmatchable keypoints.
	\end{enumerate}
\end{enumerate}

\section{Related Work}

\subsubsection{Combinatorial Optimization Meets Deep Learning}

The research on this intersection is driven by two main paradigms.

The first one attempts to improve combinatorial optimization algorithms with
deep learning methods.  Such examples include the use of reinforcement learning
for increased performance of branch-and-bound
decisions~\cite{branch-and-bound,learn-branch,gasse2019exact} as well as of
heuristic greedy algorithms for NP-Hard graph
problems~\cite{kool2018attention,tsp-policy-gradient,neural-comb-with-rl,NIPS2017_7214}.

The other mindset aims at enhancing the expressivity of neural nets by turning
combinatorial algorithms into differentiable building blocks.  The work on
differentiable quadratic programming~\cite{amos2017optnet} served as a
catalyzer and progress was achieved even in more discrete
settings~\cite{predict-and-optimize-comb,miplayer,wang2019satnet}.  In a recent
culmination of these efforts~\cite{VlastelicaEtal2020:BBoxSolvers}, a
``differentiable wrapper'' was proposed for \emph{blackbox implementations} of
algorithms minimizing a linear discrete objective, effectively allowing free
flow of progress from combinatorial optimization to deep learning.

\subsubsection{Combinatorial Graph Matching}

This problem, also known as the quadratic assignment
problem~\cite{lawler1963quadratic} in the combinatorial optimization
literature, is famous for being one of the practically most difficult
NP-complete problems.  There exist instances with less than 100 nodes that can
be extremely challenging to solve with existing
approaches~\cite{burkard1997qaplib}.  Nevertheless, in computer vision
efficient algorithmic approaches have been proposed that can routinely solve
sparse instances with hundreds of nodes.  Among those, solvers based on
Lagrangian
decomposition~\cite{GraphMatchingDDTorresaniEtAl,HungarianBP,Swoboda2016ASO}
have been shown to perform especially well, being able to quickly produce high
quality solutions with small gaps to the optimum.  Lagrange decomposition
solvers split the graph matching problem into many small subproblems linked
together via Lagrange multipliers.  These multipliers are iteratively updated
in order to reach agreement among the individual subproblems, typically with
subgradient based techniques~\cite{storvik2000lagrangian} or dual block
coordinate ascent~\cite{swoboda2017dual}.
\goodbreak

Graph matching solvers have a rich history of applications in computer vision.
A non-exhaustive list includes uses for finding correspondences of landmarks
between various objects in several semantic object
classes~\cite{GraphMatchingDDTorresaniEtAl,FactorizedGraphMatching,ufer2017deep},
for estimating sparse correspondences in wide-displacement optical
flow~\cite{GraphMatchingDDTorresaniEtAl,alhaija2015graphflow}, for establishing
associations in multiple object tracking~\cite{chen2001multi}, for object
categorization~\cite{duchenne2011graph}, and for matching cell nuclei in
biological image analysis~\cite{10.1007/978-3-319-10404-1_11}.

\subsubsection{Peer Methods}

Wider interest in deep graph matching was ignited by~\cite{zanfir2018deep}
where a fully differentiable graph matching solver based on spectral methods
was introduced.  While differentiable relaxation of quadratic graph matching
has reappeared~\cite{wang2019neural}, most
methods~\cite{wang2019learning,jiang2019glmnet,Yu2020Learning} rely on the
Sinkhorn iterative normalization~\cite{sinkhorn1967,adams2011ranking} for the
linear assignment problem or even on a single row
normalization~\cite{fey2020deep}.  Another common feature is the use of various
graph neural networks~\cite{Scarselli,li2016gated,47094} sometimes also in a
cross-graph fashion~\cite{wang2019learning} for refining the node embeddings
provided by the backbone architecture.  There has also been a discussion
regarding suitable loss
functions~\cite{zanfir2018deep,wang2019learning,Yu2020Learning}.  Recently,
nontrivial progress has been achieved by extracting more signal from the
available geometric information~\cite{fey2020deep,Zhang_2019_ICCV}.

\begin{figure}[tb]
  \centering
  \includegraphics[width=\linewidth]{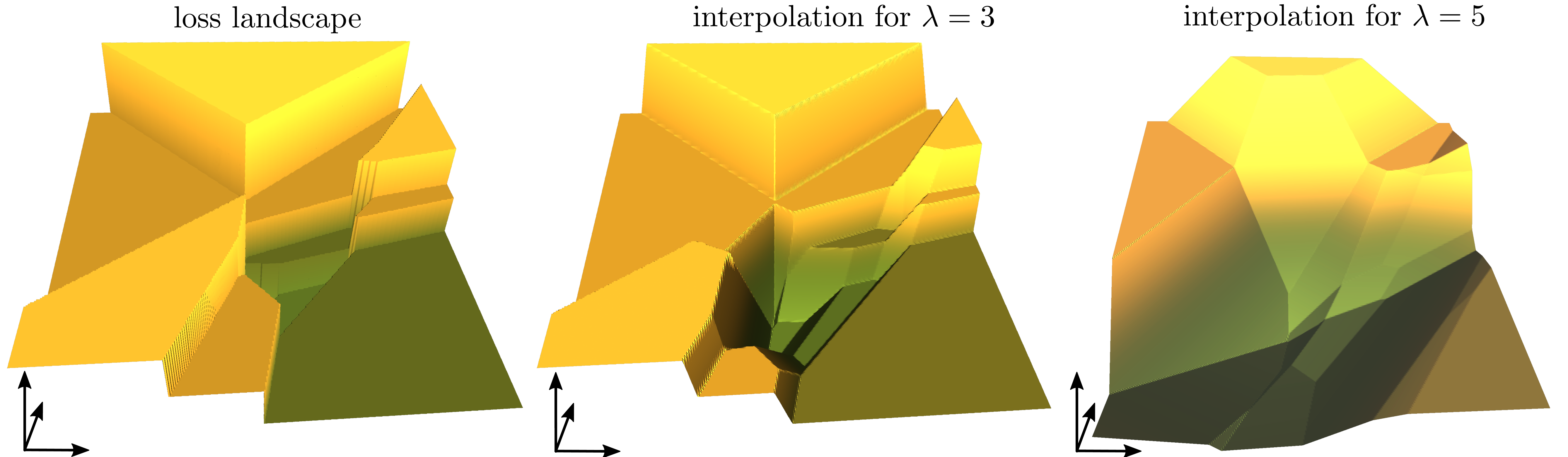}
	\caption{Differentiation of a piecewise constant loss resulting from
	incorporating a graph matching solver. A two-dimensional section of the loss
	landscape is shown (left) along with two differentiable interpolations of
	increasing strengths (middle and right).}
  \label{fig:f-lambda-2d}
\end{figure}

\section{Methods}\label{sec:methods}

\subsection{Differentiability of Combinatorial Solvers}\label{sec:bbdiff}

When incorporating a combinatorial solver into a neural network,
differentiability constitutes the principal difficulty.  Such solvers take
continuous inputs (vertex and edge costs in our case) and return a discrete
output (an indicator vector of the optimal matching).  This mapping is
piecewise constant because a small change of the costs typically does not
affect the optimal matching.  Therefore, the gradient exists almost everywhere
but is equal to zero.  This prohibits any gradient-based optimization.

A recent method proposed in~\cite{VlastelicaEtal2020:BBoxSolvers} offers a
mathematically-backed solution to overcome these obstacles.  It introduces an
efficient ``implicit interpolation'' of the solver's mapping while still
treating the solver as a blackbox.  In end effect, the intact solver is
executed on the forward pass and as it turns out, only one other call to the
solver is sufficient to provide meaningful gradient information during the
backward pass.

Specifically, the method of~\cite{VlastelicaEtal2020:BBoxSolvers} applies to
\emph{solvers} that solve an optimization problem of the form
\begin{equation} \label{E:BB-solver}
	w\in\R^N \mapsto y(w)\in Y \subset \R^N
		\quad\text{such that}\quad
		y(w) = \argmin_{y\in Y} w\cdot y,
\end{equation}
where $w$ is the continuous input and $Y$ is \emph{any} discrete set.  This
general formulation covers large classes of combinatorial algorithms that
include the shortest path problem, the traveling salesman problem and many
others.  As will be shown in the subsequent sections, graph matching is also
included in this definition.

If $L$ denotes the final loss of the network, the suggested gradient of the
piecewise constant mapping $w\mapsto L\bigl(y(w)\bigr)$ takes the form
\begin{equation} \label{E:BB-grad}
	\frac{\mathrm d L\bigl(y(w)\bigr)}{\mathrm d y}
		:= \frac{y(w_\lambda) - y(w)}{\lambda},
\end{equation}
in which $w_\lambda$ is a certain modification of the input $w$ depending on
the gradient of $L$ at $y(w)$.  This is in fact the \emph{exact gradient} of a
piecewise linear interpolation of $L\bigl(y(w)\bigr)$ in which a hyperparameter
$\lambda>0$ controls the interpolation range as \fig{f-lambda-2d} suggests.

It is worth pointing out that the framework does not require any
\emph{explicit} description of the set $Y$ (such as via linear constraints).
For further details and mathematical guarantees,
see~\cite{VlastelicaEtal2020:BBoxSolvers}.

\subsection{Graph Matching}

\newcommand{\adm}{\mathrm{Adm}(G_1, G_2)}
\newcommand{\gm}{\operatorname{GM}}

The aim of graph matching is to find an assignment between vertices
of two graphs that minimizes the sum of local and geometric costs.

Let $G_1=(V_1,E_1)$ and $G_2=(V_2,E_2)$ be two directed graphs.  We denote
by $\vv\in\{0,1\}^{|V_1||V_2|}$ the indicator vector of matched vertices,
that is $\vv_{i,j}=1$ if a vertex $i\in V_1$ is matched with $j\in V_2$ and
$\vv_{i,j}=0$ otherwise.  Analogously, we set $\ve\in\{0,1\}^{|E_1||E_2|}$ as the
indicator vector of matched edges. Obviously, the vector $\ve$ is fully
determined by the vector $\vv$. Further, we denote by $\adm$ the set of all
pairs $(\vv,\ve)$ that encode a valid matching between $G_1$ and $G_2$.

Given two cost vectors $c^v \in\R^{|V_1||V_2|}$ and $c^e\in\R^{|E_1||E_2|}$, we
formulate the graph matching optimization problem as
\begin{equation} \label{E:GM-problem}
	\gm(c^v, c^e)
		= \argmin_{(\vv, \ve)\in\adm}
			\left\{ c^v\cdot\vv + c^e\cdot\ve \right\}.
\end{equation}
It is immediate that $\gm$ fits the definition of the solver given in
\eqref{E:BB-solver}.  If $L=L(\vv, \ve)$ is the loss function, the mapping
\begin{equation} \label{E:piecewise-constant}
	(c^v, c^e) \mapsto L\bigl( \gm (c^v, c^e)\bigr)
\end{equation}
is the piecewise constant function for which the scheme
of~\cite{VlastelicaEtal2020:BBoxSolvers} suggests
\begin{equation} \label{E:GM-grad}
	\nabla\Bigl( L\bigl( \gm (c^v, c^e)\bigr)\Bigr)
		:= \frac{1}{\lambda}
			\bigl[ \gm(c^v_\lambda, c^e_\lambda) - \gm(c^v, c^e) \bigr],
\end{equation}
where the vectors $c^v_\lambda$ and $c^e_\lambda$ stand for
\begin{equation} \label{E:GM-perturbations}
	c^v_\lambda
		= c^v + \lambda \nabla_{\!\vv} L
			\bigl( \gm(c^v, c^e) \bigr)
			\quad\text{and}\quad
	c^e_\lambda
		= c^e + \lambda \nabla_{\!\ve} L
			\bigl( \gm(c^v, c^e) \bigr),
\end{equation}
where $\nabla L = (\nabla_{\!\vv} L, \nabla_{\!\ve} L)$.
The implementation is listed in \alg{main}.

In our experiments, we use the Hamming distance between the proposed matching
and the ground truth matching of vertices as a loss. In this case, $L$ does not
depend on $\ve$ and, consequently, $c^e_\lambda=c^e$.

\begin{algorithm}[tb]
\begin{subalgorithm}[t]{.48\textwidth}
	\begin{algorithmic}
		\Function{ForwardPass}{$c^v,c^e$}
		\State
		$(\vv, \ve) := \text{\bfseries GraphMatching}(c^v, c^e)$
		\\ \Comment{Run the solver}
		\Save $(\vv, \ve)$ and $(c^v, c^e)$
		\\ \Comment{Needed for backward pass}
		\Return $(\vv, \ve)$
		\EndFunction
	\end{algorithmic}
	\end{subalgorithm}%
	\begin{subalgorithm}[t]{.52\textwidth}
	\begin{algorithmic}
		\Function{BackwardPass}{$\nabla L(\vv, \ve)$, $\lambda$}
		\Load $(\vv, \ve)$ and $(c^v, c^e)$
		\State $(c^v_\lambda, c^e_\lambda)
			:= (c^v, c^e) + \lambda\nabla L(\vv,\ve)$
		\\ \Comment{Calculate modified costs}
		\State $(\vv_\lambda, \ve_\lambda)
			:= \text{\bfseries GraphMatching}(c^v_\lambda, c^e_\lambda)$
		\\ \Comment{One more call to the solver}
		\Return $\frac1\lambda\bigl(\vv_\lambda-\vv, \ve_\lambda-\ve\bigr)$
		\EndFunction
	\end{algorithmic}
\end{subalgorithm}
\caption{Forward and Backward Pass}\label{alg:main}%
\end{algorithm}

\begin{wrapfigure}{o}{.48\textwidth}
  \vspace*{-1.5\baselineskip}
  \setlength{\belowcaptionskip}{-2.\baselineskip}
	\centering
	\includegraphics[width=.97\linewidth]{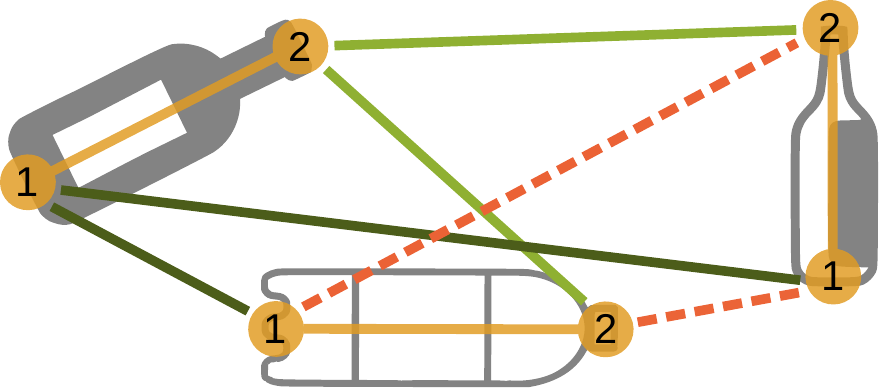}
	\caption{Cycle consistency in multi-graph matching. The partial matching
	induced by light and dark green edges prohibits including the dashed edges.}
	\label{fig:cycle-consistency}
\end{wrapfigure}

A more sophisticated variant of graph matching involves more than two graphs.  The aim
of multi-graph matching is to find a matching for every pair of graphs such
that these matchings are consistent in a global fashion (\ie satisfy so-called
cycle consistency, see \fig{cycle-consistency}) and minimize the global cost.
Although the framework of \cite{VlastelicaEtal2020:BBoxSolvers} is also
applicable to multi-graph matching, we will only use it for post-processing.

\subsection{Cost Margin}

One disadvantage of using Hamming distance as a loss function is that it
reaches its minimum value zero even if the ground truth matching has only
fractionally lower cost than competing matchings. This increases sensitivity to
distribution shifts and potentially harms generalization. The issue was already
observed in~\cite{rolinek2020cvpr}, where the
method~\cite{VlastelicaEtal2020:BBoxSolvers} was also applied. We adopt the
solution proposed in~\cite{rolinek2020cvpr}, namely the \textit{cost margin}.
In particular, during training we increase the unary costs that correspond to
the ground truth matching by $\alpha>0$, \ie
\begin{equation} \label{E:margin}
	\overleftrightarrow{c^v}_{i,j} =
	\begin{cases}
		c^v_{i,j} + \alpha &\text{if $\vv^*_{i,j}=1$}
			\\
		c^v_{i,j} & \text{if $\vv^*_{i,j}=0$}
	\end{cases}
	\quad\text{for $i\in V_1$ and $j\in V_2$},
\end{equation}
where $\vv^*$ denotes the ground truth matching indicator vector.  In all
experiments, we use $\alpha=1.0$.
\goodbreak

\subsection{Solvers}

\paragraph{Graph matching.}

We employ a dual block coordinate ascent solver~\cite{Swoboda2016ASO} based on
a Lagrange decomposition of the original problem.  In every iteration, a dual
lower bound is monotonically increased and the resulting dual costs are used to
round primal solutions using a minimum cost flow solver. We choose this solver for its state-of-the-art performance and also because it has a highly optimized publicly available implementation.

\paragraph{Multi-graph matching.}

We employ the solver from~\cite{swoboda2019convex} that builds
upon~\cite{Swoboda2016ASO} and extends it to include additional constraints
arising from cycle consistency.  Primal solutions are rounded using a special
form of permutation synchronization~\cite{pachauri2013solving} allowing for
partial matchings.

\subsection{Architecture Design}

Our end-to-end trainable architecture for keypoint matching consists of three
stages. We call it BlackBox differentiation of Graph Matching solvers
(\method).

\begin{enumerate}
\item \emph{Extraction of visual features}\hskip1ex
	A standard CNN architecture extracts a feature vector for each of the
	keypoints in the image.  Additionally, a global feature vector is extracted.
\item \emph{Geometry-aware feature refinement}\hskip1ex
	Keypoints are converted to a graph structure with spatial information.  Then
	a graph neural network architecture is applied.
\item \emph{Construction of combinatorial instance}\hskip1ex
	Vertex and edge similarities are computed using the graph features and the
	global features.  This determines a graph matching instance that is passed to
	the solver.
\end{enumerate}

\begin{figure}[b]
  \centering
  \includegraphics[width=\textwidth]{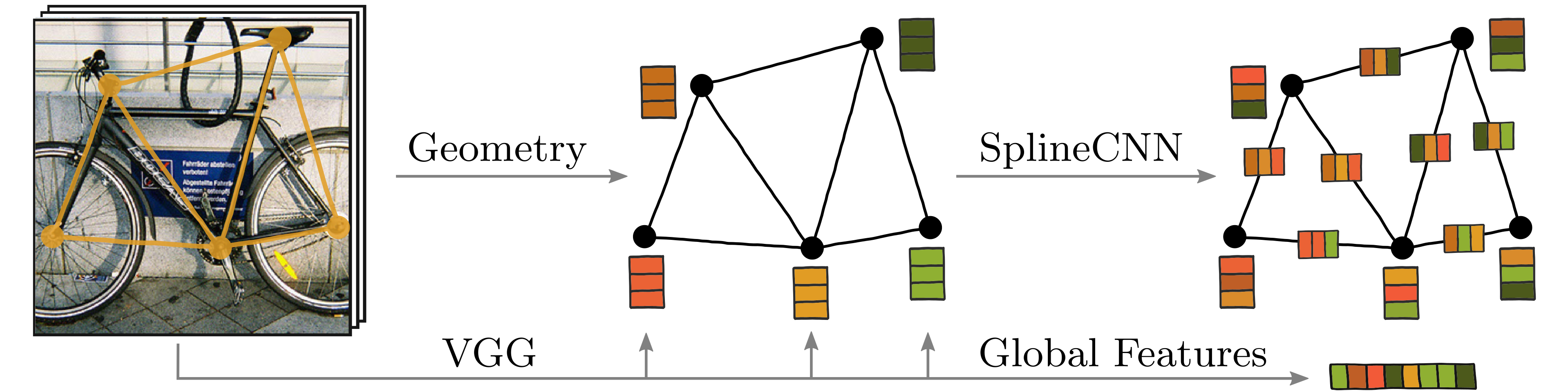}
	\caption{Extraction of features for a single image.  Keypoint locations and
	VGG features are processed by a SplineCNN and a global feature vector is
	produced.}
  \label{fig:arch-graph-features}
\end{figure}

The resulting matching $\vv$ is compared to the ground truth matching $\vv^*$
and their Hamming distance $L(\vv) = \vv \cdot (1 - \vv^*) + \vv^* \cdot (1 -
\vv)$ is the loss function to optimize.

While the first and the second stage (\fig{arch-graph-features}) are rather
standard design blocks, the third one (\fig{arch-matching}) constitutes the
principal novelty. More detailed descriptions follow.

\subsubsection{Visual Feature Extraction}

We closely follow previous work
\cite{fey2020deep,zanfir2018deep,wang2019learning} and also compute the outputs
of the \texttt{relu4\_2} and \texttt{relu5\_1} operations of the
VGG16~\cite{simonyan2014very} network pre-trained on
ImageNet~\cite{deng2009imagenet}. The spatially corresponding feature vector
for each keypoint is recovered via bi-linear interpolation.

An image-wide global feature vector is extracted by max-pooling the output of
the final VGG16 layer, see \fig{arch-graph-features}. Both the keypoint feature
vectors and the global feature vectors are normalized with respect to the $L^2$
norm.

\subsubsection{Geometric Feature Refinement}

The graph is created as a Delaunay triangulation~\cite{delaunay1934} of the
keypoint locations.
Each edge consists of a pair of directed edges pointing in opposite directions.
We deploy SplineCNN~\cite{fey2018splinecnn}, an
architecture that proved successful in point-cloud processing.  Its inputs are
the VGG vertex features and spatial edge attributes defined as normalized
relative coordinates of the associated vertices (called anisotropic
in~\cite{fey2020deep, dgmc-repo}).  We use two layers of SplineCNN with
\textsc{max} aggregations.  The outputs are additively composed with the
original VGG node features to produce the refined node features. For subsequent
computation, we set the edge features as the differences of the refined node
features.  For illustration, see \fig{arch-graph-features}.

\begin{figure}[t]
  \centering
  \includegraphics[width=\textwidth]{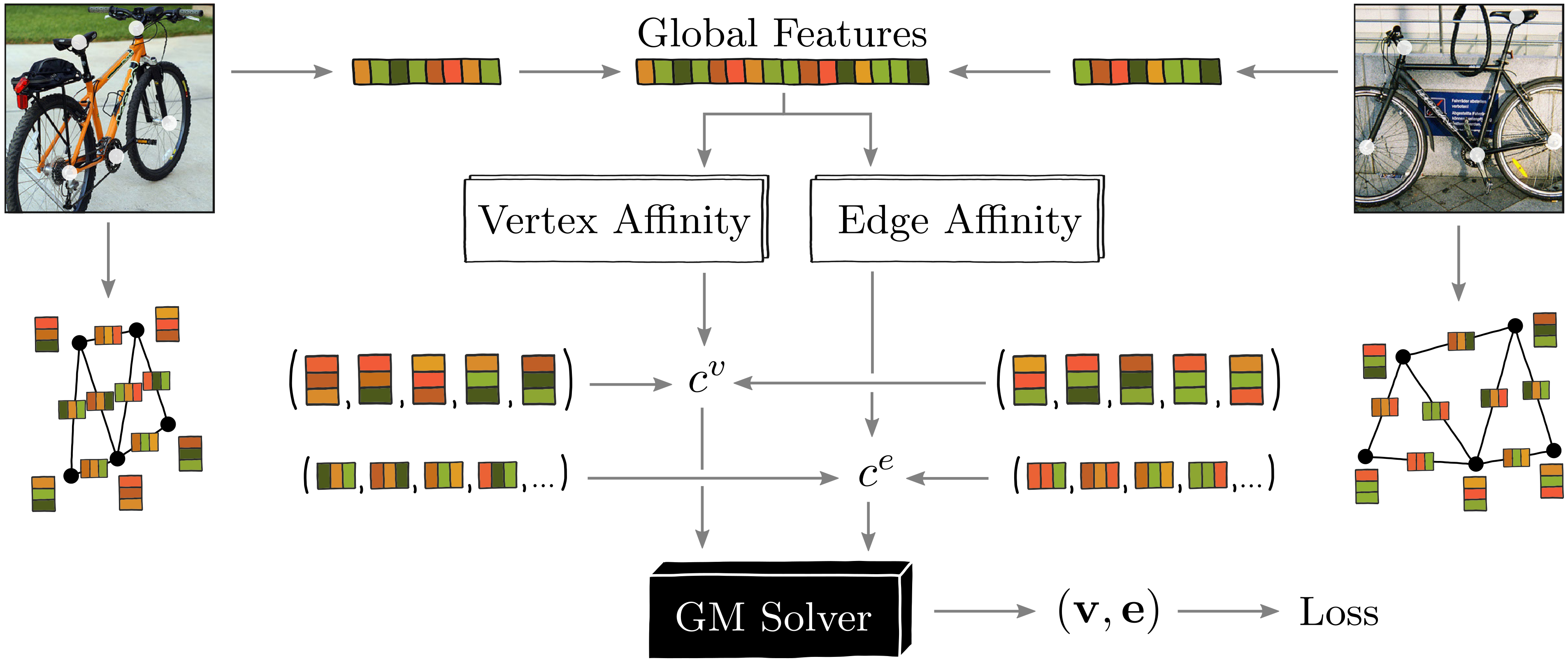}
  \caption{Construction of combinatorial instance for keypoint matching.}
  \label{fig:arch-matching}
\end{figure}

\subsubsection{Matching Instance Construction}

Both source and target image are passed through the two described procedures.
Their global features are concatenated to one global feature vector $g$.  A
standard way to prepare a matching instance (the unary costs $c^v$) is to
compute the inner product similarity (or affinity) of the vertex features
$c^v_{i,j} = f^v_{s}(i)\cdot f^v_{t}(j)$, where $f^v_s(i)$ is the feature
vector of the vertex $i$ in the source graph and $f^v_t(j)$ is the feature
vector of the vertex $j$ in the target graph, possibly with a learnable vector
or a matrix of coefficient as~in~\cite{wang2019learning}.

In our case, the vector of ``similarity coefficients'' is produced as
a weighted inner product
\begin{equation} \label{E:affinity}
	c^v_{i,j} = \sum_{k} f^v_{s}(i)_k\, a_k\, f^v_{t}(j)_k,
\end{equation}
where $a$ is produced by a one-layer NN from the global feature vector $g$.
This
allows for a gating-like behavior; the individual coordinates of the feature
vectors may play a different role depending on the global feature vector $g$.
It is intended to enable integrating various global semantic aspects such as
rigidity of the object or the viewpoint perspective.  Higher order cost terms
$c^e$ are calculated in the same vein using edge features instead of vertex
features with an analogous learnable affinity layer. For an overview, see
\fig{arch-matching}.

\section{Experiments}

We evaluate our method on the standard datasets for keypoint matching \VOC with
Berkeley annotations~\cite{everingham2010pascal,bourdev2009poselets} and
\WillowOC~\cite{Cho2013:learning}. Additionally, we propose a harder setup for
\VOC that avoids keypoint filtering as a preprocessing step. Finally, we report
our performance on a recently published dataset \SPair~\cite{min2019spair71k}.
Even though this dataset was designed for a slightly different community, its
high quality makes it very suitable also in this context. The two new
experimental setups aim to address the lack of difficult benchmarks in this
line of work.

In some cases, we report our own evaluation of DGMC \cite{fey2020deep}, the
strongest competing method, which we denote by DGMC$^*$. We used the publicly
available implementation \cite{dgmc-repo}.

\subsubsection{Runtime}

All experiments were run on a single Tesla-V100 GPU. Due to the efficient
C\texttt{++} implementation of the solver \cite{swoboda2017dual}, the
computational bottleneck of the entire architecture is evaluating the VGG
backbone. Around 30 image pairs were processed every second.

\subsubsection{Hyperparameters}

In all experiments, we use the exact same set of hyperparameters.  Only the
number of training steps is dataset-dependent.  The optimizer in use is Adam
\cite{kingma2014adam} with an initial learning rate of $2\times 10^{-3}$ which
is halved four times in regular intervals.  Learning rate for finetuning the
VGG weights is multiplied with $10^{-2}$.  We process batches of $8$ image
pairs and the hyperparameter $\lambda$ from \eqref{E:BB-grad} is consistently
set to $80.0$.  For remaining implementation details, the full code base will
be made available.

\subsubsection{Image Pair Sampling and Keypoint Filtering} \label{sec:sampling}

The standard benchmark datasets provide images with annotated keypoints but do
not define pairings of images or which keypoints should be kept for the
matching instance.  While it is the designer's choice how this is handled
during training it is imperative that only one pair-sampling and keypoint
filtering procedure is used at test time. Otherwise, the change in the
distribution of test pairs and the corresponding instances may have unintended
effects on the evaluation metric (as we demonstrate below), and therefore
hinder fair comparisons.

We briefly describe two previously used methods for creating evaluation data,
discuss their impact, and propose a third one.

\paragraph{Keypoint intersection (\interKP)}

\begin{figure}[tb]
  \centering
  \begin{subfigure}[b]{.47\textwidth}
    \setlength{\belowcaptionskip}{0pt}
    \centering
    \includegraphics[width=\linewidth]{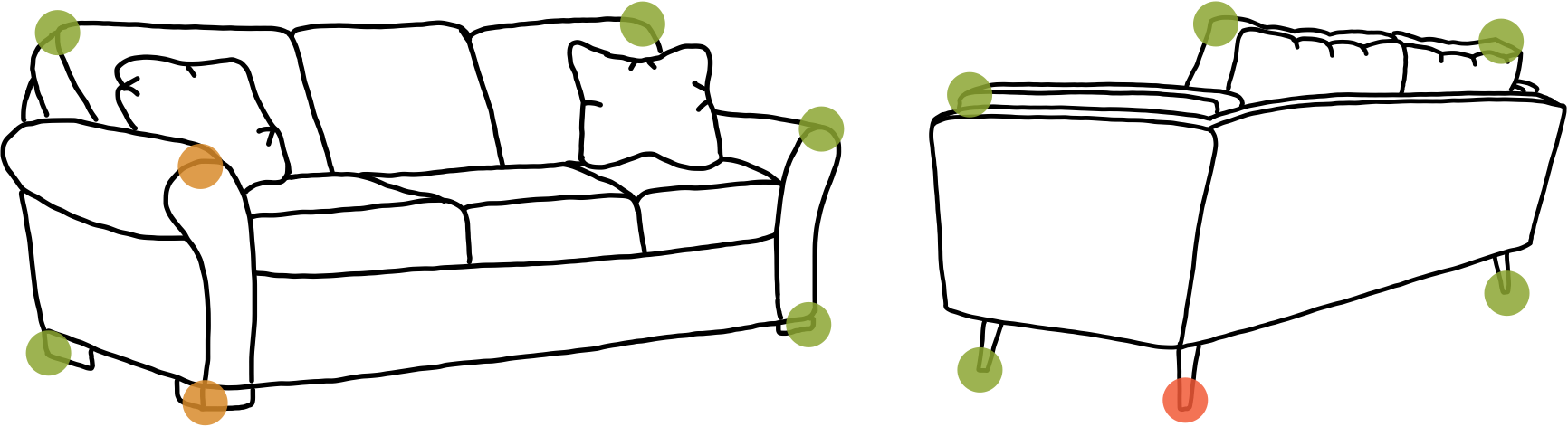}
		\caption{Intersection filtering (\interKP). Only the keypoints visible in
		both images are used (green), others are ignored (yellow, red).}
    \label{fig:sampling:square}
  \end{subfigure}
  \hskip 0pt plus 1fill
  \begin{subfigure}[b]{.47\textwidth}
    \setlength{\belowcaptionskip}{0pt}
    \centering
    \includegraphics[width=\linewidth]{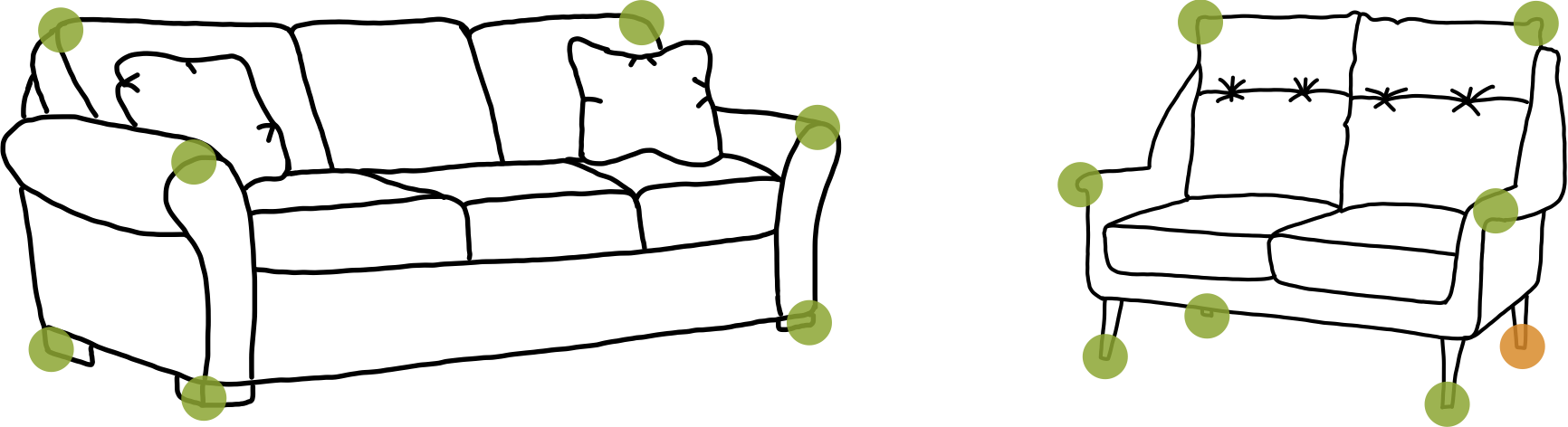}
		\caption{Inclusion filtering (\incluKP). For any source image (left), only
		the targets (right) containing all the source keypoints are used.}
    \label{fig:sampling:superset}
  \end{subfigure}
	\caption{Keypoint filtering strategies.  The image pair in (a) would not
	occur under inclusion filtering (b) because the different perspectives lead
	to incomparable sets of keypoints.  Intersection filtering is unaffected by
	viewpoints.}
  \label{fig:sampling}
\end{figure}

Only the keypoints present in both source and target image are preserved for
the matching task. In other words, all outliers are discarded. Clearly, any pair of images can be processed this way, see
\fig{sampling:square}.

\paragraph{Keypoint inclusion (\incluKP)}

Target image keypoints have to include all the source image keypoints.  The
target keypoints that are not present in the source image are then disregarded. The source image may still contain outliers.
Examples in which both target and source images contain outliers such as in \fig{sampling:superset}, will not be present.

\medskip

When keypoint inclusion filtering is used on evaluation, some image pairs are
discarded, which introduces some biases.  In particular, pairs of images seen
from different viewpoints become underrepresented, as such pairs often have
uncomparable sets of visible keypoints, see \fig{sampling}.  Another effect is
a bias towards a higher number of keypoints in a matching instance which makes
the matching task more difficult.  While the effect on mean accuracy is not
strong, \tab{PascalVOCSampling} shows large differences in individual classes.

\begin{table}[bt]
\renewcommand{\arraystretch}{1.2}
\setlength{\fHeight}{2.5ex}
\setlength{\belowcaptionskip}{-.5\baselineskip}
\centering
\footnotesize
\caption{Impact of filtering strategies on test accuracy (\%) for DGMC
\cite{fey2020deep} on \VOC.  Classes with drastic differences are highlighted.}
\begin{adjustbox}{max width=\textwidth}
	\begin{tabular}{r@{\hskip.5em} *{20}{c@{\hskip4px}} @{\hskip.5em} c <{\clearrow}}
	\toprule
	\textbf{Filter} & \tableIcons & \textbf{Mean}
		\\ \midrule
	\interKP
		& 50.4 & 67.6 & \hlcell70.7 & 70.5 & 87.2 & 85.2 & 82.5 & 74.3 & 46.2 & 69.4 & \hlcell69.9 &\hlcell 73.9 & 73.8 & 65.4 & \hlcell51.6 & 98.0 & 73.2 &\hlcell69.6 & 94.3 & 89.6 & $73.2 \pm 0.5$
		\\
	\incluKP
		& 45.5 & 66.6 & \hlcell54.5 & 67.8 & 87.2 & 86.4 & 85.6 & 73.2 & 38.5 & 67.3 & \hlcell86.9 & \hlcell64.9 & 78.9 & 60.3 & \hlcell61.5 & 96.8 & 68.7 & \hlcell93.5 & 93.6 & 85.0 & $73.1 \pm 0.4$
		\\
	\bottomrule
	\end{tabular}
\end{adjustbox}
\label{tab:PascalVOCSampling}
\end{table}

Another unsatisfactory aspect of both methods is that label information is
required at evaluation time, rendering the setting quite unrealistic.  For this
reason, we \textbf{propose to evaluate without any keypoint removal}.

\paragraph{Unfiltered keypoints (\unfilKP)}

For a given pair of images, the keypoints are used without any filtering.
Matching instances may contain a different number of source and target
vertices, as well as outliers in both images. This is the most general setup.

\subsection{\VOC}

The \VOC~\cite{everingham2010pascal} dataset with Berkeley annotations
\cite{bourdev2009poselets} contains images with bounding boxes surrounding
objects of 20 classes.  We follow the standard data preparation procedure
of~\cite{wang2019learning}.  Each object is cropped to its bounding box and
scaled to $256\times 256\,\mathrm{px}$.  The resulting images contain up to 23
annotated keypoints, depending on the object category.

The results under the most common experimental conditions (\interKP) are
reported in \tab{PascalVOCBig} and we can see that \method outperforms
competing approaches.

\begin{table}[tb]
\renewcommand{\arraystretch}{1.2}
\setlength{\fHeight}{2.5ex}
\footnotesize
\centering
\caption{Keypoint matching accuracy (\%) on \VOC using standard intersection
filtering (\interKP).  For GMN~\cite{zanfir2018deep} we report the improved
results from~\cite{wang2019learning} denoted as GMN-PL.  DGMC*
is~\cite{fey2020deep} reproduced using \interKP.  For DGMC* and \method  we
report the mean over 5 restarts.}
\begin{adjustbox}{max width=\textwidth}
	\begin{tabular}{r@{\hskip1em} *{20}{c} @{\hskip1em} c <{\clearrow}}
	\toprule
	\textbf{Method} & \tableIcons & \textbf{Mean}
		\\ \midrule
	GMN-PL
		& 31.1 & 46.2 & 58.2 & 45.9 & 70.6 & 76.5 & 61.2 & 61.7 & 35.5 & 53.7
		& 58.9 & 57.5 & 56.9 & 49.3 & 34.1 & 77.5 & 57.1 & 53.6 & 83.2 & 88.6
		& 57.9
		\\
	PCA-GM~\cite{wang2019learning}
		& 40.9 & 55.0 & 65.8 & 47.9 & 76.9 & 77.9 & 63.5 & 67.4 & 33.7 & 66.5
		& 63.6 & 61.3 & 58.9 & 62.8 & 44.9 & 77.5 & 67.4 & 57.5 & 86.7 & 90.9
		& 63.8
		\\
	NGM+~\cite{wang2019neural}
		& 50.8 & 64.5 & 59.5 & 57.6 & 79.4 & 76.9 & 74.4 & 69.9 & 41.5 & 62.3
		& 68.5 & 62.2 & 62.4 & 64.7 & 47.8 & 78.7 & 66.0 & 63.3 & 81.4 & 89.6
		& 66.1
		\\
	GLMNet~\cite{jiang2019glmnet}
		& 52.0 & 67.3 & 63.2 & 57.4 & 80.3 & 74.6 & 70.0 & 72.6 & 38.9 & 66.3
		& 77.3 & 65.7 & 67.9 & 64.2 & 44.8 & 86.3 & 69.0 & 61.9 & 79.3 & 91.3
		& 67.5
		\\
	CIE$_1$-H~\cite{Yu2020Learning}
		& 51.2 & 69.2 & 70.1 & 55.0 & 82.8 & 72.8 & 69.0 & 74.2 & 39.6 & 68.8
		& 71.8 & 70.0 & 71.8 & 66.8 & 44.8 & 85.2 & 69.9 & 65.4 & 85.2 & 92.4
		& 68.9
		\\
	DGMC$^*$~\cite{fey2020deep}
    & 50.4 & 67.6 & 70.7 & 70.5 & 87.2 & 85.2 & 82.5 & 74.3 & 46.2 & 69.4
		& 69.9 & 73.9 & 73.8 & 65.4 & 51.6 & \textbf{98.0} & 73.2 & 69.6 & 94.3 & 89.6
		& $73.2 \pm 0.5$
		\\
	\method
		& \textbf{61.5} & \textbf{75.0} & \textbf{78.1} & \textbf{80.0} & \textbf{87.4} & \textbf{93.0} & \textbf{89.1} & \textbf{80.2} & \textbf{58.1}
		& \textbf{77.6} & \textbf{76.5} & \textbf{79.3} & \textbf{78.6} & \textbf{78.8} & \textbf{66.7} & 97.4 & \textbf{76.4} & \textbf{77.5} & \textbf{97.7} & \textbf{94.4}
		& $\mathbf{80.1\pm0.6}$
		\\
	\bottomrule
	\end{tabular}
\end{adjustbox}
\label{tab:PascalVOCBig}
\end{table}

\begin{table}[bt]
\renewcommand{\arraystretch}{1.2}
\setlength{\fHeight}{2.5ex}
\footnotesize
\centering
\caption{F1 score (\%) for \VOC keypoint matching without filtering (\unfilKP).
	As a reference we report an ablation of our method where the solver is forced
	to match all source keypoints, denoted as \method{}-Max.  \method-Multi refers
	to using the multi-graph matching solver with cycle
	consistency~\cite{swoboda2019convex} with sets of 5 images at evaluation.  The
	reported statistics are over 10 restarts.  The last row displays the percentage
	of unmatched keypoints in the test-set pairs.}
\begin{adjustbox}{max width=\textwidth}
	\begin{tabular}{r@{\hskip.5em} *{20}{c} @{\hskip.5em} c <{\clearrow}}
	\toprule
	\textbf{Method} & \tableIcons & \textbf{Mean}
		\\\midrule
	\method{}-Max
		& 35.5 & 68.6 & 46.7 & 36.1 & 85.4 & 58.1 & 25.6 & 51.7 & 27.3
		& 51.0 & 46.0 & 46.7 & 48.9 & 58.9 & 29.6 & 93.6 & 42.6 & 35.3  & 70.7 &  79.5
		& $51.9\pm1.0$
		\\
	\method
		& 42.7 & \textbf{70.9} & 57.5 & 46.6 & \textbf{85.8} & 64.1 & 51.0 & 63.8 & 42.4
		& 63.7 & 47.9 & 61.5 & \textbf{63.4} & 69.0 & \textbf{46.1} & 94.2 & \textbf{57.4} & 39.0  & \textbf{78.0} &  82.7
		& $61.4\pm0.5$
		\\
	\method-Multi
	    & \textbf{43.4} & 70.5 & \textbf{61.9} & \textbf{46.8} & 84.9 & \textbf{65.3} & \textbf{54.2} & \textbf{66.9} & \textbf{44.9}
	    & \textbf{67.5} & \textbf{50.8} & \textbf{66.8} & 63.3 & \textbf{71.0} & \textbf{46.1} & \textbf{96.1} & 56.5 & \textbf{41.3}  & 73.4 & \textbf{83.4}
	    & $\mathbf{62.8 \pm 0.5}$
	    \\ \midrule
	  \setrow{\itshape} Unmatched (\%)
		& 22.7 & 4.9 & 30.6 & 29.1 & 2.7 & 23.8 & 40.8 & 26.4 & 17.3
		& 25.1 & 21.2 & 27.4 & 26.8 & 16.6 & 22.1 & 6.7 & 36.7 & 27.5  & 31.7 &  14.0
		& \it 22.7         \\
         \bottomrule
	\end{tabular}
\end{adjustbox}
\label{tab:PascalVOCunfiltered}
\end{table}

\subsubsection{All keypoints}

We propose, see \sect{sampling}, to preserve all keypoints (\unfilKP).
Matching accuracy is no longer a good evaluation metric as it ignores false
positives.  Instead, we report F1-Score, the harmonic mean of precision and
recall.

Since the underlying solver used by our method also works for partial
matchings, our architecture is applicable out of the box.  Competing
architectures rely on either the Sinkhorn normalization or a softmax and as
such, they are hard-wired to produce maximal matchings and do not offer a
simple adjustment to the unfiltered setup.  To simulate the negative impact of
maximal matchings we provide an ablation of \method where we modify the solver
to output maximal matchings. This is denoted by  \method{}-Max.

In addition, we report the scores obtained by running the multi-graph matching
solver \cite{swoboda2019convex} as post-processing. Instead of sampling pairs
of images, we sample sets of 5 images and recover from the architecture the
costs of the $\binom{5}{2} = 10$ matching instances. The multi-graph matching
solver then searches for globally optimal set of consistent matchings.  The
results are provided in \tab{PascalVOCunfiltered}.

Note that sampling sets of 5 images instead of image pairs does not interfere
with the statistics of the test set. The results are therefore comparable.

\subsection{\WillowOC}

The \WillowOC dataset contains a total of 256 images from 5 categories. Each
category is represented by at least 40 images, all of them with consistent
orientation.  Each image is annotated with the same 10 distinctive
category-specific keypoints, which means there is no difference between the
described keypoint filtering methods.  Following standard procedure, we crop
the images to the bounding boxes of the objects and rescale to $256\times
256\,\mathrm{px}$.

Multiple training strategies have been used in prior work.  Some authors decide
to train only on the relatively small Willow dataset, or pretrain on \VOC and
fine-tune on Willow afterward~\cite{wang2019learning}.  Another approach is to
pretrain on \VOC and evaluate on Willow without fine-tuning, to test the
transfer-ability~\cite{wang2019neural}.  We report results for all different
variants, following the standard procedure of using 20 images per class when
training on Willow and excluding the classes \emph{car} and \emph{motorbike}
from \VOC when pre-training, as these images overlap with the Willow dataset.
We also evaluated the strongest competing approach DGMC \cite{fey2020deep}
under all settings.

\begin{table}[t]
\centering
\footnotesize
\caption{Keypoint matching accuracy (\%) on \WillowOC.  The columns Pt and Wt
	indicate training on \VOC and Willow, respectively.  Comparisons should be made
	only within the same training setting.  For HARG-SSVM~\cite{Cho2013:learning}
	we report the comparable figures from \cite{wang2019learning}. Twenty restarts
	were carried out.}
\begin{adjustbox}{max width=\textwidth}
\setrow{\textbf}%
\begin{tabular}{R@{\hskip1em}*{7}{C}<{\clearrow}}
	\toprule
	Method & Pt & Wt & face & motorbike & car & duck & bottle
		\\ \midrule
	HARG-SSVM~\cite{wang2019learning}\setrow{\hlcell}
	& x & \checkmark & 91.2 & 44.4 & 58.4 & 55.2 & 66.6
		\\ \cmidrule{2-8}
	{\multirow{2}{*}{GMN-PL~\cite{wang2019learning,zanfir2018deep}}}
		\setrow{\hhlcell}
	& \checkmark & x					& 98.1 & 65.0 & 72.9 & 74.3 & 70.5
		\\
	& \checkmark & \checkmark & 99.3 & 71.4 & 74.3 & 82.8 & 76.7
		\\ \cmidrule{2-8}
	{\multirow{2}{*}{PCA-GM~\cite{wang2019learning}}}
		\setrow{\hhlcell}
	& \checkmark & x & 100.0 & 69.8 & 78.6 & 82.4 & 95.1
		\\
	& \checkmark & \checkmark & 100.0 & 76.7 & 84.0 & 93.5 & 96.9
		\\ \cmidrule{2-8}
	{\multirow{2}{*}{CIE~\cite{Yu2020Learning}}}
		\setrow{\hhlcell}
	& \checkmark & x					& 99.9 & 71.5 & 75.4 & 73.2 & 97.6
		\\
	& \checkmark & \checkmark & 100.0 & 90.0 & 82.2 & 81.2 & 97.6
		\\ \cmidrule{2-8}
	NGM~\cite{wang2019neural}\setrow{\hlcell}
	& x					 & \checkmark & 99.2 & 82.1 & 84.1 & 77.4 & 93.5
		\\ \cmidrule{2-8}
	GLMNet~\cite{jiang2019glmnet}
	& \checkmark & \checkmark & 100.0 & 89.7 & 93.6 & 85.4 & 93.4
		\\ \cmidrule{2-8}
	{\multirow{3}{*}{\makecell{DGMC*~\cite{fey2020deep}}}}
		\setrow{\hhlcell}
	& \checkmark & x					& $98.6\pm1.1$ & $69.8\pm 5.0$ & $84.6\pm 5.2$ & $76.8\pm 4.3$ & $90.7\pm 2.4$
		\\ \setrow{\hlcell}
	& x					 & \checkmark & $100.0\pm 0.0$ & $98.5\pm 1.5$ &	$98.3\pm 1.2$ & $90.2\pm 3.6$ & $98.1\pm 0.9$
		\\
	& \checkmark & \checkmark & $100.0\pm 0.0$ & $98.8 \pm 1.6$ & $96.5\pm 1.6$ & $93.2\pm 3.8$ & $99.9\pm 0.3$
		\\ \cmidrule{2-8}
	{\multirow{3}{*}{\method}}
		\setrow{\hhlcell}
	& \checkmark & x					& $100.0\pm 0.0$ & $95.8\pm 1.4$ & $89.1\pm 1.7$ & $89.8\pm 1.7$ & $97.9\pm 0.7$
		\\ \setrow{\hlcell}
	& x					 & \checkmark & $100.0\pm 0.0$ & $99.2\pm 0.4$ & $96.9\pm 0.6$ & $89.0\pm 1.0$ & $98.8\pm 0.6$
		\\
	& \checkmark & \checkmark & $100.0 \pm 0.0$ & $98.9 \pm 0.5$ & $95.7 \pm 1.5$ & $93.1 \pm 1.5$ & $99.1 \pm 0.4$
		\\
	\bottomrule
\end{tabular}
\end{adjustbox}
\label{tab:Willow}
\end{table}

The results are shown in \tab{Willow}.  While our method achieves good
performance, we are reluctant to claim superiority over prior work. The small
dataset size, the multitude of training setups, and high standard deviations
all prevent statistically significant comparisons.

\subsection{\SPair}

We also report performance on \SPair~\cite{min2019spair71k}, a dataset recently
published in the context of dense image matching. It contains 70,\,958 image
pairs prepared from \VOC 2012 and Pascal 3D+. It has several advantages over
the \VOC dataset, namely higher image quality, richer keypoint annotations,
difficulty annotation of image-pairs, as well as the removal of the ambiguous
and poorly annotated \textit{sofas} and \textit{dining tables}.

\begin{figure}[h]
  \setlength{\belowcaptionskip}{-0.5\baselineskip}
  \includegraphics[align=t,width=0.32\linewidth]{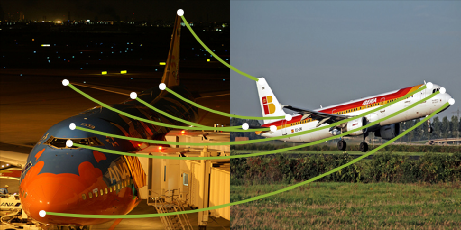}
	\hfil
  \includegraphics[align=t,width=0.32\linewidth]{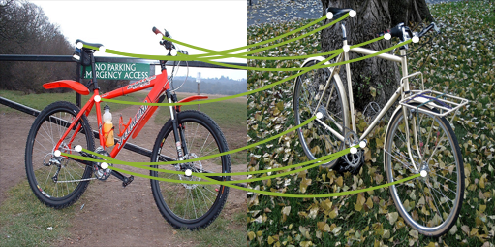}
	\hfil
  \includegraphics[align=t,width=0.32\linewidth]{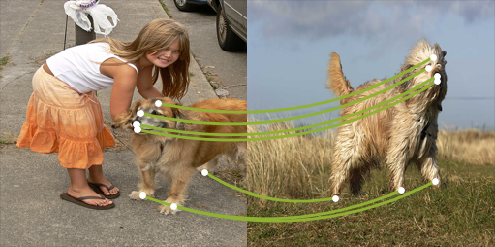}
	\\ [4pt]
  \includegraphics[align=t,width=0.32\linewidth]{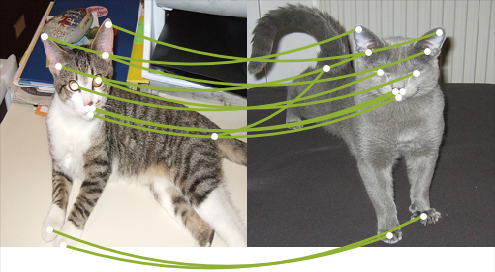}
	\hfil
  \includegraphics[align=t,width=0.32\linewidth]{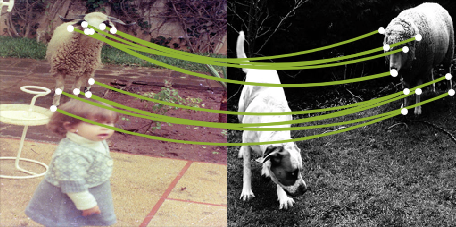}
	\hfil
  \includegraphics[align=t,width=0.32\linewidth]{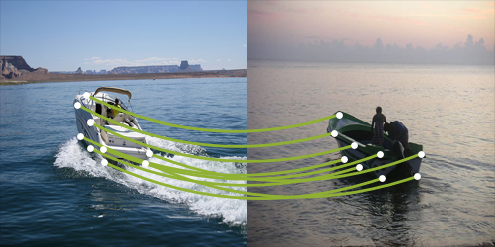}
	\\
  \includegraphics[align=t,width=0.32\linewidth]{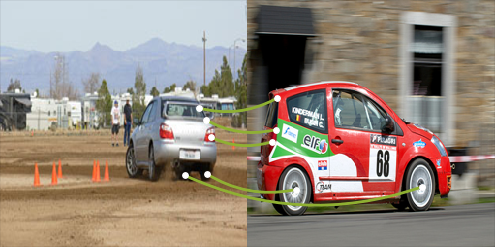}
	\hfil
  \includegraphics[align=t,width=0.32\linewidth]{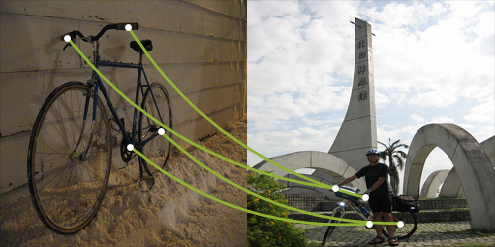}
	\hfil
  \includegraphics[align=t,width=0.32\linewidth]{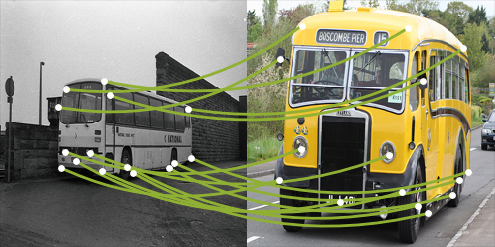}
	\\ [4pt]
  \includegraphics[align=t,width=0.32\linewidth]{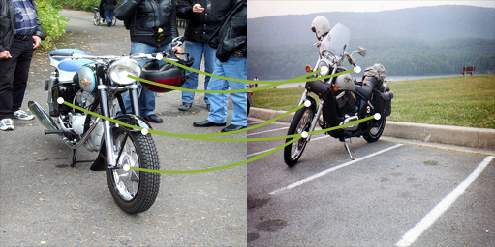}
	\hfil
  \includegraphics[align=t,width=0.32\linewidth]{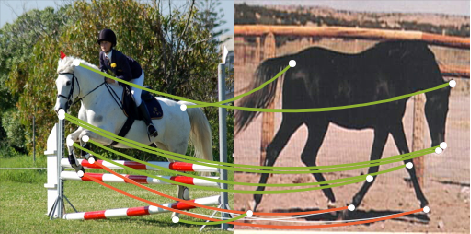}
	\hfil
  \includegraphics[align=t,width=0.32\linewidth]{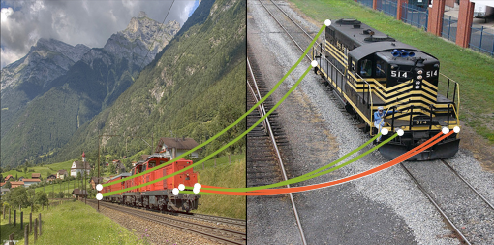}
	\\ [4pt]
  \includegraphics[align=t,width=0.32\linewidth]{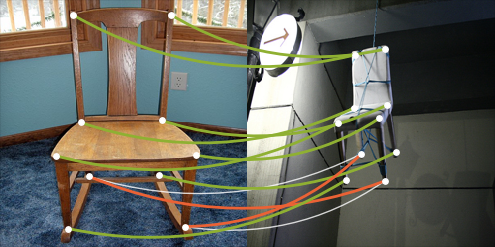}
	\hfil
  \includegraphics[align=t,width=0.32\linewidth]{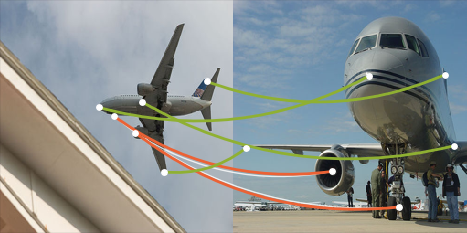}
	\hfil
  \includegraphics[align=t,width=0.32\linewidth]{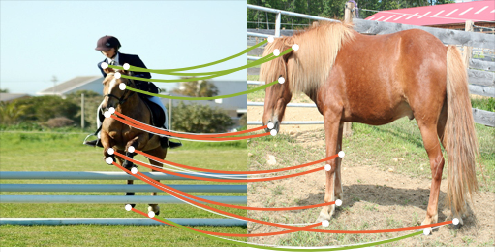}
  \caption{Example matchings from the \SPair dataset.
  }
  \label{fig:examples2}
\end{figure}

Again, we evaluated DGMC \cite{fey2020deep} as the strongest competitor of our
method. The results are reported in \tab{Spair} and \tab{Spair-classes}. We
consistently improve upon the baseline, particularly on pairs of images seen
from very different viewpoints. This highlights the ability of our method to
resolve instances with conflicting evidence. Some example matchings are
presented in \fig{examples} and \fig{examples2}.

\begin{table}
\caption{Keypoint matching accuracy (\%) on \SPair grouped by levels of
	difficulty in the viewpoint of the matching-pair. Statistics is over 5
	restarts.}
\centering
\scriptsize
\begin{adjustbox}{max width=\textwidth}
\begin{tabular}{l@{\hskip1em}c@{\hskip1em}c@{\hskip1em}c@{\hskip1em}c}
	\toprule
	\multirow{2}{*}{\bfseries Method} & \multicolumn{3}{c}{\bfseries Viewpoint difficulty} & \multirow{2}{*}{\bfseries All}
		\\
	& \textbf{easy} & \textbf{medium} & \textbf{hard} &
		\\ \midrule
	DGMC*& $79.4 \pm 0.2$ & $65.2 \pm 0.2$ & $61.3 \pm 0.5$ & $72.2\pm0.2$
		\\
	\method & $\mathbf{84.8 \pm 0.1}$ & $\mathbf{73.1 \pm 0.2}$ & $\mathbf{70.6 \pm 0.9}$ & $\mathbf{78.9\pm 0.4}$
		\\
	\bottomrule
\end{tabular}
\end{adjustbox}
\label{tab:Spair}
\end{table}

\begin{table}
\setlength{\fHeight}{2.5ex}
\footnotesize
\renewcommand{\tableIcons}{%
\faPlane & \faBicycle & \faCrow & \faShip & \faWineBottle & \faBus &
\faCar & \faCat &	\faChair & \faCow & \faDog & \faHorse &
\faMotorcycle & \faWalking & \faTulip& \faSheep & \faTrain & \faTv}
\renewcommand{\arraystretch}{1.2}
\centering
\caption{Keypoint matching accuracy (\%) on \SPair for all classes.}
\begin{adjustbox}{max width=\textwidth}
	\setrow{\bfseries}
	\begin{tabular}{r@{\hskip1em}*{18}{c}@{\hskip1em}c<{\clearrow}}
	\toprule
	Method & \tableIcons & Mean
		\\ \midrule
	DGMC*
		& 54.8 & 44.8 & 80.3 & 70.9 & 65.5 & 90.1 & 78.5 & 66.7 & 66.4
		& 73.2 & 66.2 & 66.5 & 65.7 & 59.1 & 98.7 & 68.5 & 84.9 & 98.0
		& $72.2 \pm0.2$
		\\
	\method
	\setrow{\bfseries}
		& 66.9 & 57.7 & 85.8 & 78.5 & 66.9 & 95.4 & 86.1 & 74.6 & 68.3
		& 78.9 & 73.0 & 67.5 & 79.3 & 73.0 & 99.1 & 74.8 & 95.0 & 98.6
		& $\mathbf{78.9\pm0.4}$
		\\
	\bottomrule
	\end{tabular}
\end{adjustbox}
\label{tab:Spair-classes}
\end{table}

\subsection{Ablations Studies}
To isolate the impact of single components of our architecture, we conduct
various ablation studies as detailed in the supplementary material.
The results on \VOC are summarized in Tab.~S1.

\section{Conclusion}

We have demonstrated that deep learning architectures that integrate
combinatorial graph matching solvers perform well on deep graph matching
benchmarks.

Opportunities for future work now fall into multiple categories. For one, it
should be tested whether such architectures can be useful outside the
designated playground for deep graph matching methods. If more progress is
needed, two major directions lend themselves: (i) improving the neural network
architecture even further so that input costs to the matching problem become
more discriminative and (ii) employing better solvers that improve in terms of
obtained solution quality and ability to handle a more complicated and
expressive cost structure (\eg hypergraph matching solvers).

Finally, the potential of building architectures around solvers for other
computer vision related combinatorial problems such as \textsc{multicut} or
\textsc{max-cut} can be explored.

\newpage
\begin{center}
\Large\bfseries
Supplementary Material
\end{center}
\vskip .8cm

\section*{Ablation Studies} \label{sec:abblations}

To isolate the impact of single components of our architecture, we conduct
various ablation studies.  The results on \VOC are summarized in
\tab{PascalVOCAblation} where large performance differences are highlighted.

\paragraph{Global Features / Fixed Affinity}

Influence of the global feature vector is removed. In \eqref{E:affinity},
we use a single learnable vector $a$ instead of one-layer NN of $g$.

\paragraph{VGG Fine-Tuning / Frozen VGG}

VGG pretrained on ImageNet~\cite{deng2009imagenet} is used without fine-tuning.
This ablation is in particular important for a fair comparison with
DGMC~\cite{fey2020deep} where not fine-tuning is applied as well.

\paragraph{Quadratic Costs / Unary Costs}

The matching instances consist only of unary costs $c^v$. The quadratic costs
$c^e$ are set to zero.

\paragraph{Comparison to Sinkhorn}

We use Sinkhorn normalization~\cite{sinkhorn1967,adams2011ranking} instead of
the solver.  We only obtained good performance after increasing an internal
constant $\varepsilon$ (intended to prevent division by zero) (from $10^{-8}$
to $10^{-4}$). We believe this is connected to issues with ``vanishing
gradient'' that were also reported in \cite{fey2020deep,zhang2018learning}.

\begin{table}[tbhp]
\setlength{\belowcaptionskip}{-0.5\baselineskip}
\renewcommand{\arraystretch}{1.2}
\setlength{\fHeight}{2.5ex}
\centering
\caption{Ablations of \method on \VOC.  Large performance differences are
	highlighted. Statistics is over 5 restarts.}
\begin{adjustbox}{max width=\textwidth}
	\begin{tabular}{r@{\hskip.7em} *{20}{c@{\hskip4px}} @{\hskip.3em} c <{\clearrow}}
	\toprule
	\textbf{Ablation} & \tableIcons & \textbf{Mean}
		\\ \midrule
	Unmodified
		& 62.0 & 77.0 & 76.4 & 75.9 & 89.1 & 93.7 & 88.6 & 80.0 & 56.6 & 78.2
		& 81.3 & 79.0 & 77.1 & 77.6 & 64.3 & 97.0 & 78.3 & 78.4 & 97.9 & 94.6
		& $80.1\pm0.3$
		\\
	Fixed Affinity
		& 59.1 & \hlcell 73.9 & 75.6 & 73.7 & 87.2 & \hlcell 88.4 & 86.4 & 77.7 & 55.2 & 76.2
		& \hlcell 74.7 & 77.4 & 78.1 & 77.1 & 62.7 & 96.3 & \hlcell 75.2 & \hlcell 73.1 & 96.3 & 94.4
		& $77.9\pm0.7$
		\\
	Frozen VGG
		& \hlcell 57.8 & \hlcell 73.5 & 75.2 & 77.5 & 87.2 & 92.4 & 87.1 & 79.0 & 59.1 & 77.8
		& \hlcell 77.4 & 77.9 & 77.0 & 77.4 & 63.0 & 96.9 & 75.8 & \hlcell 72.7 & 97.5 & 94.6
		& $78.9\pm0.5$
		\\
	Unary Costs
		& 60.7 & 74.4 & 77.2 & 78.0 & 87.0 & 92.3 & 89.6 & 80.4 & 55.5 & 77.3
		& \hlcell 65.2 & 79.0 & 79.2 & 77.2 & 63.0 & 97.0 & 75.8 & \hlcell 73.3 & 96.0 & 93.6
		& $78.6\pm0.4$
		\\
	Sinkhorn
		& 62.5 & \hlcell 73.9 & 74.3 & 75.3 & 88.3 & 94.3 & 88.9 & \hlcell 76.9 & \hlcell 51.6 & 75.8
		& \hlcell 68.8 & 77.0 & 75.3 & 78.3 & \hlcell 60.2 & 97.7 & 75.8 & \hlcell 71.3 & 97.8 & 94.6
		& $77.9\pm0.5$
		\\
	\bottomrule
	\end{tabular}
\end{adjustbox}
\label{tab:PascalVOCAblation}
\end{table}


\end{document}